\documentclass[letterpaper, 10 pt, conference]{ieeeconf}  

\IEEEoverridecommandlockouts                              

\usepackage{url}

\usepackage{graphics} 
\usepackage{epsfig} 
\usepackage{mathptmx} 
\usepackage{times} 
\usepackage{amsmath} 
\usepackage{amssymb}  
\usepackage[hang]{subfigure} 
\usepackage{cite}
\usepackage{siunitx}
\usepackage{soul}
\usepackage{color}
\usepackage{xcolor}

\soulregister\ref7  
\soulregister\cite7 
\soulregister\eqref7
\soulregister\prettyref7

\usepackage{multirow}
\usepackage{algorithm} 
\usepackage{algpseudocode}
\usepackage{makecell}
\usepackage{booktabs}
\usepackage{threeparttable}

\bstctlcite{IEEEexample:BSTcontrol}

\usepackage{prettyref}
\newrefformat{fig}{Fig.~\ref{#1}}
\newrefformat{sec}{Sec.~\ref{#1}}
\newrefformat{tab}{Tab.~\ref{#1}}
\newrefformat{algo}{Algo.~\ref{#1}}
\newrefformat{eq}{(\ref{#1})}

\begin{document}

\title{\LARGE \bf
Understanding Particles From Video:\\ Property Estimation of Granular Materials via Visuo-Haptic Learning}
\author{Zeqing Zhang$^{1,2}$, Guangze Zheng$^{1,2}$, Xuebo Ji$^{1,2}$, Guanqi Chen$^{1,2}$, Ruixing Jia$^{1,2}$, Wentao Chen$^{1,2}$\\
Guanhua Chen$^{3}$, Liangjun Zhang$^{4}$ and Jia Pan$^{1,2}$%
\thanks{$^{1}$The University of Hong Kong, Hong Kong.}
\thanks{$^{2}$Center for Transformative Garment Production, Hong Kong.}
\thanks{$^{3}$Southern University of Science and Technology, China.}%
\thanks{$^{4}$Robotics and Autonomous Driving Lab, Baidu Research, USA.}
\thanks{Corresponding author: Jia Pan {\tt\small jpan@cs.hku.hk}}
}

\maketitle
\thispagestyle{empty}
\pagestyle{empty}

\begin{abstract}
Granular materials (GMs) are ubiquitous in daily life. 
Understanding their properties is also important, especially in agriculture and industry. However, existing works require dedicated measurement equipment and also need large human efforts to handle a large number of particles. In this paper, we introduce a method for estimating the relative values of particle size and density from the video of the interaction with GMs. It is trained on a visuo-haptic learning framework inspired by a contact model, which reveals the strong correlation between GM properties and the visual-haptic data during the probe-dragging in the GMs. After training, the network can map the visual modality well to the haptic signal and implicitly characterize the relative distribution of particle properties in its latent embeddings, as interpreted in that contact model.
Therefore, we can analyze GM properties using the trained encoder, and only visual information is needed without extra sensory modalities and human efforts for labeling.
The presented GM property estimator has been extensively validated via comparison and ablation experiments. The generalization capability has also been evaluated and a real-world application on the beach is also demonstrated.
{Experiment videos are available at} \url{https://sites.google.com/view/gmwork/vhlearning}.
\end{abstract}

\IEEEpeerreviewmaketitle

\section{Introduction}
\label{sec:intro}
Granular materials (GMs) are very common in daily life, such as grains in agriculture, sands in geology, etc. They are a collection of discrete solid particles, characterized by the fluidization \cite{zik1992mobility} and jamming \cite{zhang2024haptic} under external forces, separating from solids, liquids, and gases. It is important to understand the properties of GM in practice. For example, in agricultural applications, the degree of maturity can be determined by sampling the size and density of grains \cite{yang2021assessment}. In geology, the water content of the sandy environment is normally measured to determine the degree of soil softness, and then analyze the geological risk of debris flow \cite{rasheed2022soil}.

Currently, property analysis of GMs requires special tools, such as balances, vernier calipers, hygrometers, or some sophisticated and expensive devices, such as infrared sensors \cite{castilla2023thermal} or remote sensing satellites \cite{zhong2024risk}. In recent years, several works \cite{matl2020inferring, guo2023estimating} have introduced artificial intelligence into the property estimation for GMs. However, the granule itself is composed of a large number of solid particles and measurement of particle properties requires a lot of manpower. In addition, although the existing learning-based methods can perform well in their respective tasks, their interpretability has been questioned, which challenges their generalization.

\begin{figure}[!tb]
    \centering
    \includegraphics[width=0.485\textwidth]{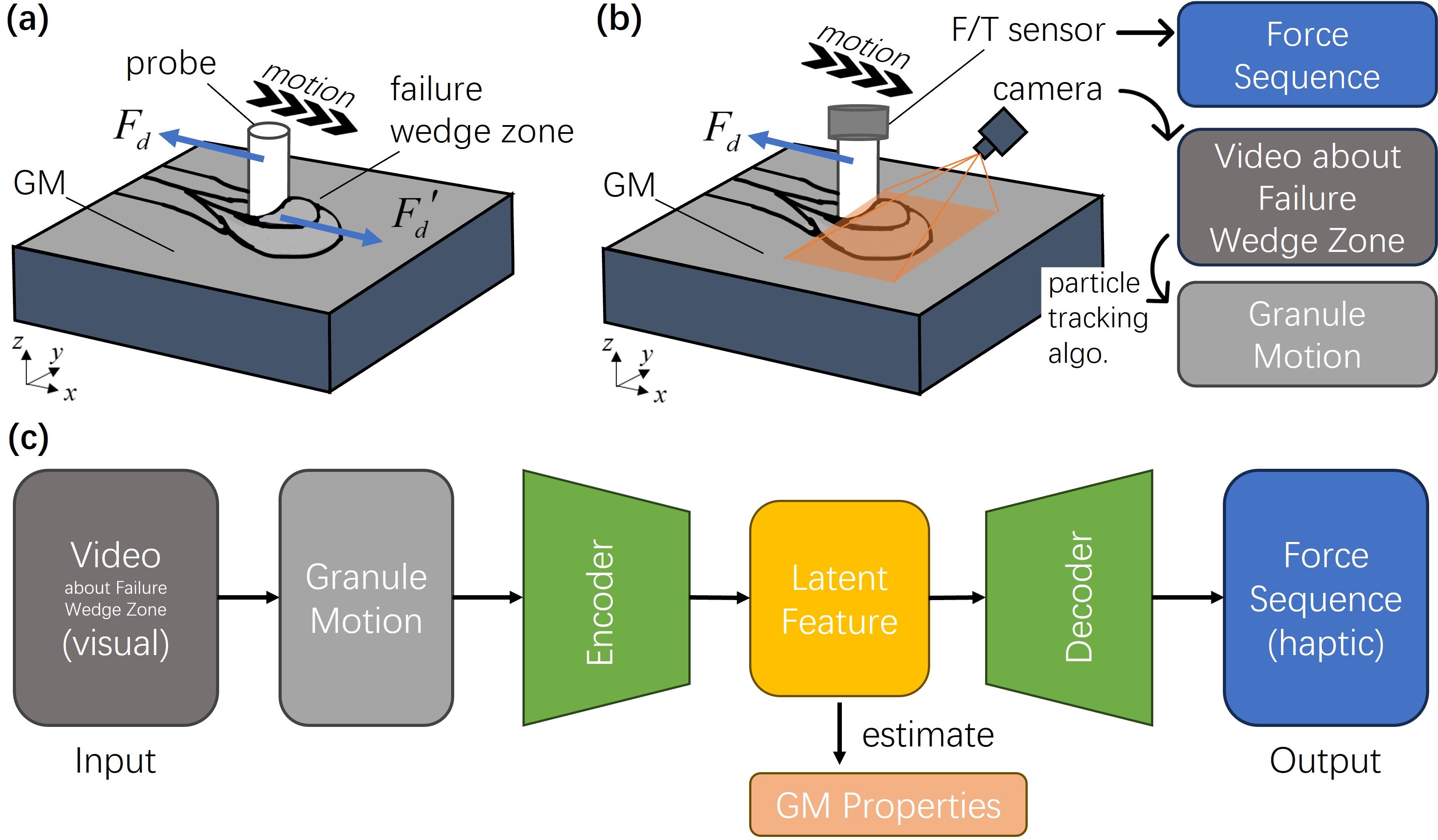}
    \vspace{-20pt}
    \caption{Overview of this work. (a) Probe-dragging. A simplified GM-tool contact model is given in the physics community \cite{albert1999slow}. (b) Visual and haptic data. Force sequence $F_d$ is measured by the F/T sensor, and the granule motion is extracted by the proposed particle tracking algorithm from a video clip. (c) Workflow of our visuo-haptic learning, where the granule properties are analyzed from the latent features after training.}
    \label{fig:first_fig}
    \vspace{-15pt}
\end{figure}

To this end, based on a GM-tool contact model \cite{albert1999slow} studied by physicists as in \prettyref{fig:first_fig}-(a), this work leverages easily-acquired video and force signals (see \prettyref{fig:first_fig}-(b)) to train an encoder-decoder network in a supervised manner, as depicted in \prettyref{fig:first_fig}-(c). By doing so, it discovers the model's understanding of granule properties in its latent embeddings, as explained in that physical contact model. Therefore, utilizing the trained encoder, this paper provides an effective visual estimator that allows users to analyze GM attributes from a video sequence alone. It significantly improves the usability and generalization of our method.


Specifically, the contact model in \cite{albert1999slow} reveals the strong correlation between visual-haptic properties (GM size and density) and particle motions, as well as contact forces, during probe drag. It inspires us to design a visuo-haptic learning architecture and employ the video of particle motions and the force sequence as the input and output, respectively.
Guided by the contact model, we successfully explore the implicit property distribution in the latent representations.
In this way, only a camera and a force sensor are necessary to obtain the training data. During training, our pipeline uses force sequences as supervisory signals, avoiding large human efforts to obtain external labels. In addition, we utilize a particle tracking algorithm modified from a pre-trained model \cite{cotracker} to further extract the motion features of the granules from raw videos, which significantly reduces the dimension of the input data and speeds up the training process. In the evaluation stage, we discard the decoder and instead leverage the encoder, which maps the video of probe-dragging in certain GM to the implicit property distribution in the latent embeddings. Based on the projection position in this property distribution, the relative values of particle size and density of this kind of GM can be estimated.
Note that the inference process does not require any additional sensors and simply requires a camera to capture the video, that is, ``understanding particles from video''. To the best of the authors' knowledge, it is the first study to investigate property estimation for GMs merely from visual modality.


\noindent{\textbf{Main contributions}}: 
\begin{itemize}
    \item We propose a method to estimate the relative property values of the GM from videos. This approach provides a useful tool in practical applications where dedicated measurement instruments are lacking.
    \item We present a visuo-haptic learning framework inspired by a contact model in GMs, using readily available visual and haptic signals. It avoids the human labeling and enhances the interpretability of the latent features.
    \item We validate the performance and generalization capabilities of the proposed property estimator. Also, its application in a real-world outdoor scenario is demonstrated.
\end{itemize}

The rest of the paper is organized as follows. \prettyref{sec:related} reviews the related work and \prettyref{sec:methodology} introduces the proposed method. \prettyref{sec:exp} demonstrates the implementation details, and extensive experiments of the proposed property estimator are conducted in \prettyref{sec:results}. Finally, the conclusion, limitation, and future work are discussed in \prettyref{sec:conclusion}.

\section{Related Work}
\label{sec:related}

\begin{figure*}
    \centering
    \includegraphics[width=0.95\textwidth]{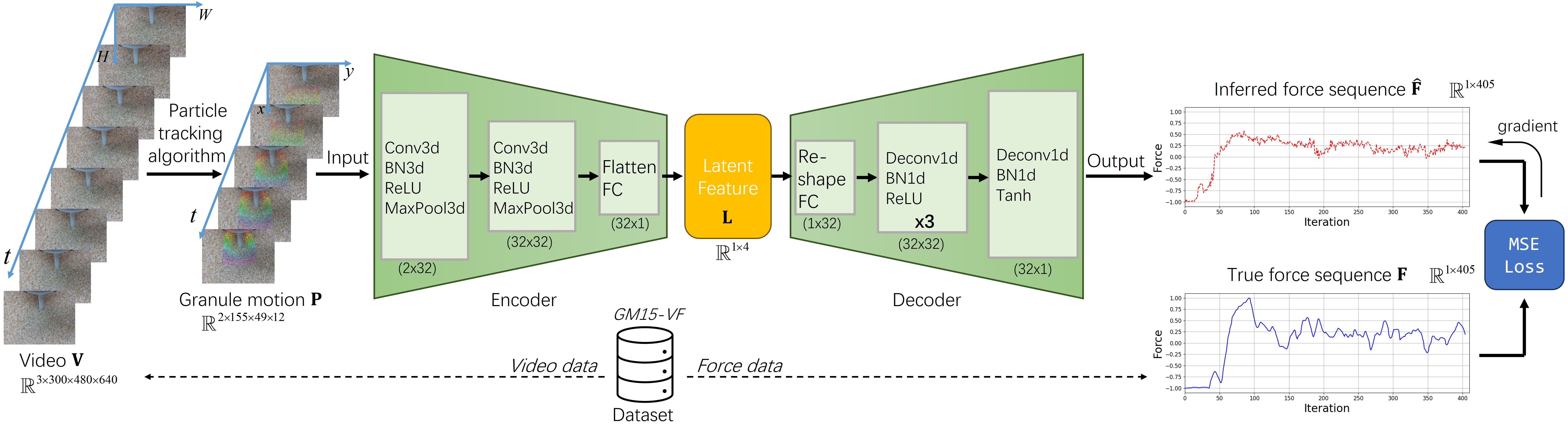}
    \vspace{-10pt}
    \caption{Architecture of our visuo-haptic learning framework inspired by the contact model in e.q. \prettyref{eq:third_law}. The dataset \textit{GM15-VF} provides the video $\mathbf{V}$ ($C3, D300, H480, W640$) about the probe-dragging and corresponding force sequence $\mathbf{F}$ ($C 1\times D 405$). After being processed by the proposed particle tracking algorithm, the encoder takes as input the trajectories $\mathbf{P}$ (in $x$ and $y$ coordinates) of $49\times12$ points throughout $155$ frames. After processing the decoder, an inferred force sequence $\hat{\mathbf{F}}$ is generated and subsequently utilized to calculate the MSE loss with the true force value $\mathbf{F}$.}
    \label{fig:framework_mae}
    \vspace{-15pt}
\end{figure*}

Modeling deformable bodies presents inherent challenges compared to rigid bodies \cite{yang2024one}.  
The precise property representations are helpful for the modeling of deformable objects and benefit the downstream manipulation tasks \cite{yang2020rigid,liu2022robot}.  

Several works take advantage of machine learning techniques to characterize properties of common deformable objects, such as liquids and cloth. For example, \cite{huang2022understanding} leverages dynamic tactile sensing to estimate the liquid volume and sugar water solution. Similar work can be found in a recent study \cite{niu2023goats} about target water-mass estimation by curriculum reinforcement learning. \cite{takahashi2019deep} proposes a fabric-related property estimator from the image via visuo-tactile learning. Also, \cite{longhini2023edo} learns graph dynamics for cloth-like objects to obtain a latent representation of elastic physical properties. 
As for GMs, since the total mass/weight is easier to obtain as a supervisory signal compared to other GM properties, the majority of research focuses on the mass-estimation task based on different sensing modalities, e.g., using audio \cite{clarke2018learning}, RGBD \cite{takahashi2021uncertainty}, force signals \cite{kiyokawa2019generation}. A recent work reports a sim-to-real power weighing policy from simulation \cite{kadokawa2023learning}.

Only a few works pay attention to the particle property estimation. \cite{matl2020inferring}
infers granular simulation parameters from the macroscopic behavior of GMs. This work requires the discrete element method (DEM) to simulate particle behavior and evaluate GM property inference, which inevitably contains the sim-to-real gap. In addition, the calibrated parameters are more specific to the given simulator, rather than the more general particle property values. 
\cite{guo2023estimating} leverages dynamic haptic sensing to estimate four particle properties from real granules enclosed in the container. 
However, in the inference stage, the customized haptic sensor, robot arm, and extra efforts to capsule GMs in the bottle are needed, limiting its application in practice. In addition, to estimate absolute property values, above methods requires external labels provided by humans as supervisory signals. 
In this paper, we design a visuo-haptic learning network based on the physical model and interpret the implicit property distribution from the latent embeddings. 
Our work collects videos and force signals in real GMs without simulated data or human efforts to get true labels. In the evaluation process, only a clip of video easily captured by the camera, e.g., from a smartphone, is needed. So it allows us to exploit it in real-world scenarios without the need for specific sensors or bulky robot arms.


\section{Methodology}
\label{sec:methodology}
\subsection{Contact Model of Probe-Dragging}
The goal of this work is to estimate the physical properties of particles directly from the visual modality. However, we do not leverage visual data and property labels to train an end-to-end network, as that is the black-box model and also requires external human efforts for labels. Instead, we focus on a well-studied scenario in the physics community, namely the probe-dragging in GMs \cite{maladen2009undulatory, zhu2019data, kobayakawa2018local}, as shown in \prettyref{fig:first_fig}-(a). In this case, when a probe is dragged through a homogeneous granular medium, a failure wedge zone \cite{swick1988model} forms in front of the probe. The drag force $F_d$ experienced by the probe arises from resistive forces among particles within this failure wedge zone. Physicists in \cite{albert1999slow} have provided a simplified contact model that reveals the explicit relationship between the drag force $F_d$ and the particle size $d_c$ and density $\rho$, expressed as
$F_d = \eta \rho g d_c H^2$,
where $\eta$ refers to the surface morphology, and $H$ is the penetration depth of the probe, and $g$ is the gravity constant.
Furthermore, the reaction force $F^\prime_d$, displayed in \prettyref{fig:first_fig}-(a), drives the displacement of all particles within the failure wedge zone, as represented by
$F^\prime_d = \sum_{i} m_i \ddot{x}_i$,
where $m_i$ and $\ddot{x}_i$ are the mass and acceleration of the particle $i$ in the failure wedge zone, respectively.
Particles outside this region remain stationary. {According to Newton's third law, if only considering the force magnitude and ignoring the force direction, we have $F_d = F^\prime_d$, that is,}
\begin{equation}\label{eq:third_law}
    \sum_{i} m_i \ddot{x}_i = \eta \rho g d_c H^2 = F_d.
\end{equation}
From it, we observe a strong correlation between GM properties (i.e., particle size $d_c$ and density $\rho$) and the particle motion $\ddot{x}_i$ as well as the drag force $F_d$. Note that both particle motion information and force signals in the drag process are readily available, as they can be directly measured using the camera and force sensor, as depicted in \prettyref{fig:first_fig}-(b).

Inspired by the contact model above, we propose a visuo-haptic learning network (see \prettyref{fig:framework_mae}), mapping the visual features of granule motions (related to $\ddot{x}_i$) to the force sequence (related to $F_d$), i.e., from left to right in \prettyref{eq:third_law}. It is due to the simultaneous integration of visual and haptic modalities that we discover an implicit property distribution about GM visual-haptic properties i.e., particle size  $d_c$ and density $\rho$, embedded in the latent representations. Note that, here we are not trying to obtain absolute values for particle properties, but just a relative distribution of their properties. Given a new GM, its size and density can be relatively estimated based on its projection location in the implicit property distribution.

\subsection{Particle Tracking Algorithm}
The input of our method is the video about the failure wedge zone captured by the camera. 
Instead of feeding high-dimension video data into the network, we employ a particle tracking algorithm based on a pre-trained tracking model \cite{cotracker} to further extract the temporal motion features of granules from the video. The particle tracking preprocessing not only reduces the input dimension, which is beneficial for model training, but the subsequent ablation studies (\prettyref{sec:ablation}) have also validated its efficacy in enhancing the estimation of GM properties.


In detail, we first delimit a region of interest (ROI) at the first frame $I_1$ of the video $\mathbf{V}$ with $T_v$ frames and then discretize the region into $m$ rows and $n$ columns of points.
These sampled points would be attached to granules shown in $I_1$, denoted as $\mathbf{P}_1 = \{P^i_t: (x^i_t, y^i_t) \in \mathbb{R}^{2}, t=1, i=1,\dots, m \times n\}$. Subsequently, a convolutional neural network (CNN) $\phi$ is employed to extract image feature $\phi(I_t) \in \mathbb{R}^{d\times h \times w}$, where $t = 1, \dots, T_v$, and $d$, $h$, and $w$ represent the dimension, height, and width of the feature, respectively. So, the individual characteristic $Q_t^i \in \mathbb{R}^{d}$ for the point $P_t^i$ can be obtained from $\phi(I_t)$ according to the location of the point in $I_t$. Features $\mathbf{Q}_{t}$ of all points $\mathbf{P}_{t}$ are then input into a pre-trained tracking model $\Psi$ \cite{cotracker}, then the positions of these points in subsequent frames are obtained to form granule motion trajectories, i.e.,
$
  \mathbf{P}_{t+1}, \mathbf{Q}_{t+1}
  = \Psi(
    \mathbf{P}_{t}, \mathbf{Q}_{t}
  )
$.
Due to the probe sliding in the GM, new granules may emerge from the bottom of the ROI in the next frame. So, we periodically sample a new row of points at the bottom of the region of interest (ROI) at a frame-wise interval and start tracking their trajectories to enable continuous acquisition of GM motions. If a sampled point exits the field of view in the video, its trajectory is considered terminated. 
{More details and results can be found on the project site given in the abstract.} 

\subsection{Encoder-Decoder Network}
We utilize the encoder-decoder network to perform visuo-haptic learning, mapping the granule motion signal to the force sequence, as illustrated in \prettyref{fig:framework_mae}.
%
%
After particle tracking preprocessing, the motion of the granules is obtained in the form of trajectories $\mathbf{P} = \{P^i_t: (x^i_t, y^i_t), t = 1, \dots, T_p, i=1,\dots, (m+\Delta m) \times n\} \in \mathbb{R}^{2\times T_p\times (m+\Delta m)\times n}$ from a video $\mathbf{V}$. Note that, $T_p < T_v$, since we only track granule motions during the constant velocity translation phase of the probe. Since we periodically augment the sampling points at the bottom of ROI, we ultimately obtain trajectory information for a matrix of $(m+\Delta m)$ rows and $n$ columns of points.
Then $\mathbf{P}$ is sent to the encoder, where we employ 3D convolutional layers to extract features. The first convolutional layer involves a 3D CNN to increase the number of channels from $2$ to $C_e$, and the second layer performs 3D convolution while maintaining the $C_e$ channels. Subsequently, a fully connected (FC) layer is employed to flatten the resulting data to the $l$-dimension latent space $\mathbf{L} \in \mathbb{R}^{1\times l}$. Then, in the decoding stage, the encoded latent feature is reshaped into $C_d$ channels through a FC layer, followed by four 1D deconvolution layers. Finally, it outputs a sequence for the estimated force $\hat{\mathbf{F}} = \{F: f_t,~t = 1, \dots, T_f \} \in \mathbb{R}^{1\times T_f}$. Then, the mean squared error (MSE) loss is calculated by comparing the $\hat{\mathbf{F}}$ with the true force sequence $\mathbf{F} \in \mathbb{R}^{1\times T_f}$ measured by F/T sensor during the probe-dragging. 



By using readily available force information as the supervisory signal, rather than relying on property labels, we have avoided the need for extensive manual labeling efforts. In addition, we further leverage a particle tracking algorithm to extract the motion features of particles from the input videos, thereby reducing the dimensionality of the input and accelerating the training process. Furthermore, since our training framework is built upon a physics-based contact model in GMs, this approach increases the interpretability of the model in the latent embeddings, as it is able to capture both visual and haptic features simultaneously.

\section{Experiments}
\label{sec:exp}
\subsection{Dataset and Experiment Setup}
\begin{figure}[!tb]
    \centering
    \includegraphics[width=0.485\textwidth]{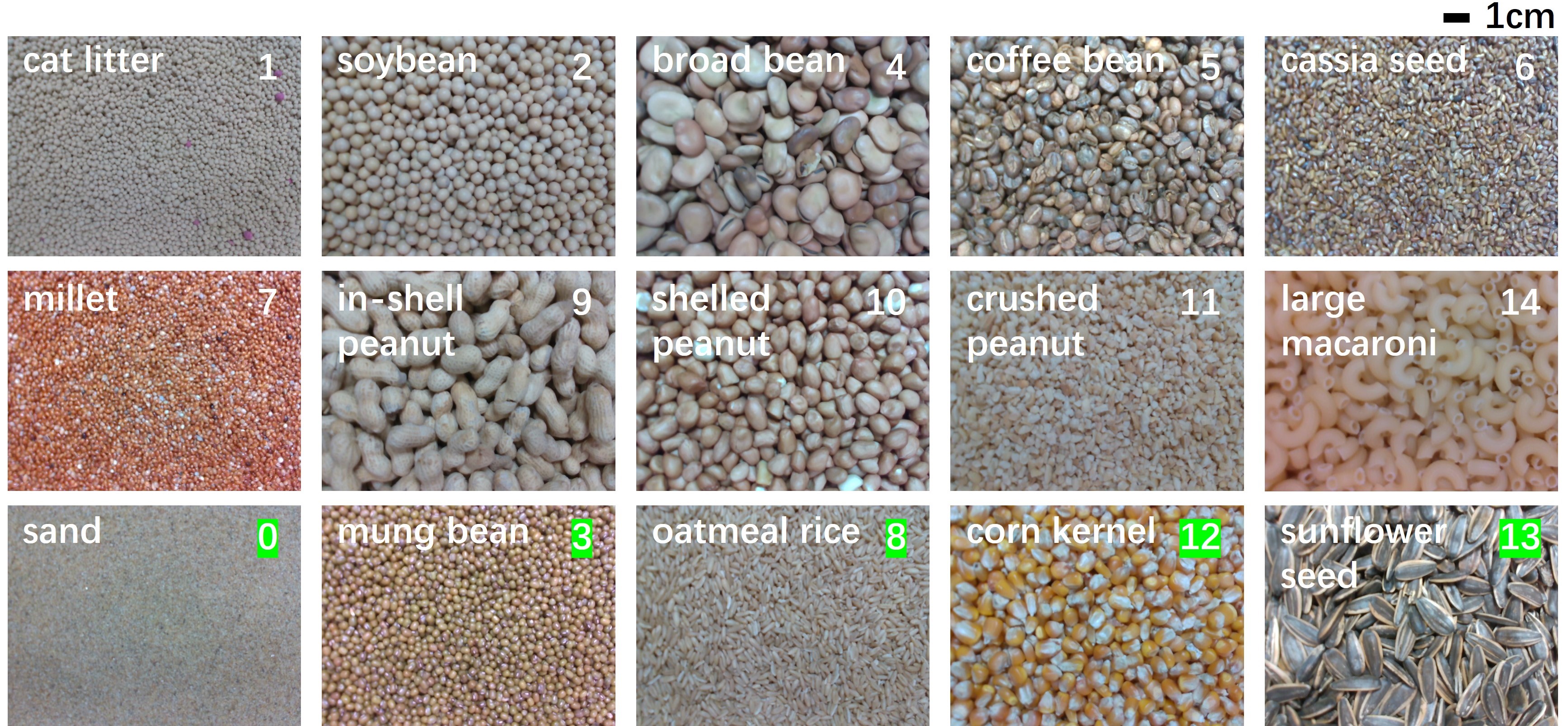}
    \vspace{-20pt}
    \caption{$15$ types of GMs compose the dataset \textit{GM15-VF}, where the unseen particles are displayed with green backgrounds for their IDs.}
    \label{fig:all_gm_fig}
    \vspace{-5pt}
\end{figure}

\begin{figure}
    \centering
    \includegraphics[width=0.485\textwidth]{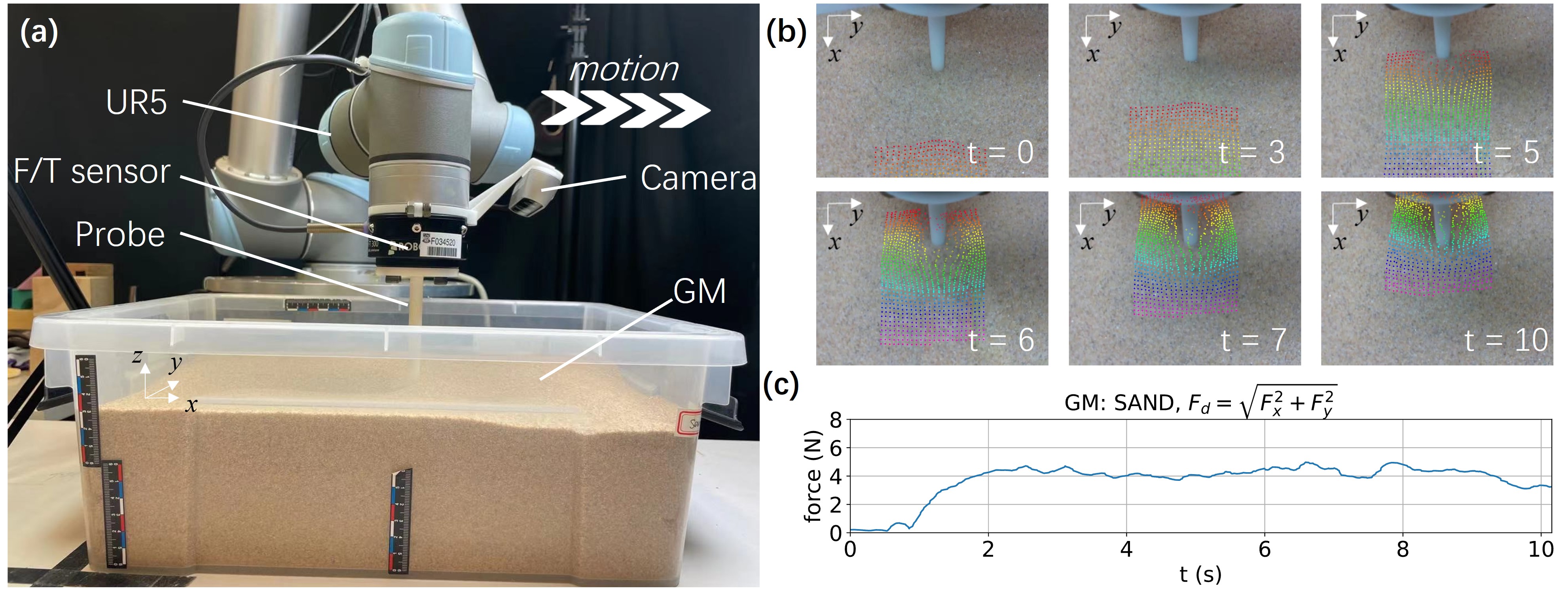}
    \vspace{-20pt}
    \caption{Data collection. (a) Experiment setup. (b) Visual data. The proposed particle tracking algorithm is employed on the video clip captured by the mounted camera. (c) Haptic data. The force $F_d$ exerted on the probe is measured by the mounted F/T sensor. Here we only consider the resultant force on the $x-y$ plane.}
    \label{fig:exp_setup_visuo_haptic_data}
    \vspace{-15pt}
\end{figure}

In this study, we have collected $15$ common GMs, as depicted in \prettyref{fig:all_gm_fig}. In each GM, we capture $100$ instances of probe-dragging videos with corresponding force signals, thereby generating our \textit{GM15-VF} dataset. 
To construct it, we set up a data acquisition system, as shown in \prettyref{fig:exp_setup_visuo_haptic_data}-(a). The system comprises a UR5 robot arm with a 6-axis force-torque sensor, i.e., F/T300 (Robotiq, Canada) attached to its end-effector, capable of capturing forces exerted on the 3D-printed ABS probe as it slides through granules. Here we only record the resultant force on the $x-y$ plane, as depicted in \prettyref{fig:exp_setup_visuo_haptic_data}-(c). Also, there is a holder at the end of the robot arm, which securely holds an Intel RealSense D435 camera to capture the variations of particles within the failure wedge zone, as shown in \prettyref{fig:exp_setup_visuo_haptic_data}-(b). For each sliding, the probe will linearly slide in GM along the $x$-axis for $14$ cm and {with a depth of $6$ cm}. {The probe motion velocity is kept at the low speed around $0.015$ m/s}.
Among our dataset \textit{GM15-VF}, we select 5 types of granules as unseen materials (last row in \prettyref{fig:all_gm_fig}) and use the remaining 10 GMs for training with $80\%$ training data, $10\%$ validation data and $10\%$ testing data, respectively.

\subsection{Implementation Details}
As depicted in \prettyref{fig:framework_mae}, we uniformly extract the $300$-frame video segment (i.e., $T_v = 300$) for each probe-dragging video, then apply the particle tracking algorithm to obtain the motion trajectories of $49 \times 12$ points tracked over $155$ frames, i.e., $ m + \Delta m = 49, n = 12, T_p = 155$. The channel dimension of both encoder and decoder is set to $C_e = C_d = 32$. The dimension of latent features is determined to be $l = 4$ based on our preliminary experiment, shown in \prettyref{sec:dim_latent}. The output dimension of the force sequence is $T_f = 405$. To training stability, we uniformly normalize the force sequence $\mathbf{F}$ and $\hat{\mathbf{F}}$ to  $[-1, 1]$.
During the training, we employ Adam \cite{kingma2014adam} as the optimizer and the learning rate is given as $1e^{-3}$ with the batch size of $32$. All codes are built on PyTorch and trained on one NVIDIA GeForce RTX 4090 GPU. 

\section{Results}\label{sec:results}
We will present results of the model evaluation, as well as the baseline experiment, ablation experiments, and generalization experiments in this section. A real-world application of the proposed estimator is also demonstrated. 




\subsection{Force Inference} \label{sec:force_infer}

\begin{figure}
    \centering
    \includegraphics[width=0.48\textwidth]{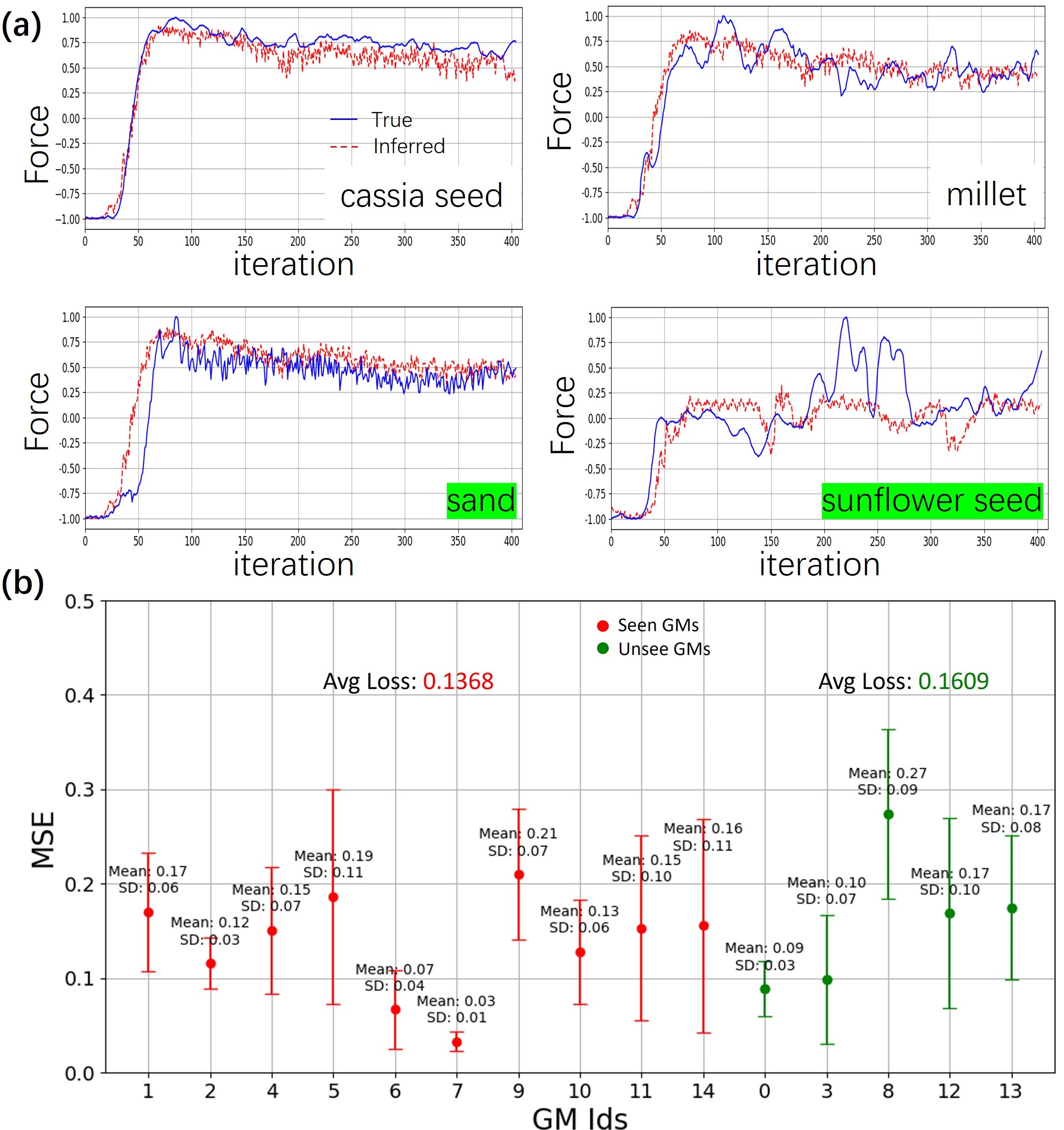}
    \vspace{-10pt}
    \caption{Force inference. (a) Predicted force sequences from inputting videos, including seen and unseen (green background) materials. (b) Means and standard deviations of MSE of force sequence for each GM.}
    \label{fig:force_inference}
    \vspace{-15pt}
\end{figure}

After training, we input the videos from the testing set into the trained network and evaluate the force inference performance by calculating the MSE between its outputs and the true force sequences corresponding to the input video. Some examples are exhibited in \prettyref{fig:force_inference}-(a), where we can observe the trained network can map the visual modality (i.e., video) to the haptic modality (i.e., force). In addition, compared to the seen particles, our model performs quite well on the unseen granules, predicting the rough force trends just from the given videos. The means and standard deviations of the MSE of force inference for all GMs in the test set are presented in \prettyref{fig:force_inference}-(b). In general, the variance of error from particles with large sizes is intended to be larger than that of small particles, such as the GM 7 (millet) v.s. GM 14 (large macaroni).

\subsection{Property Estimation} \label{sec:property_esti}
\begin{figure}
    \centering
    \includegraphics[width=0.485\textwidth]{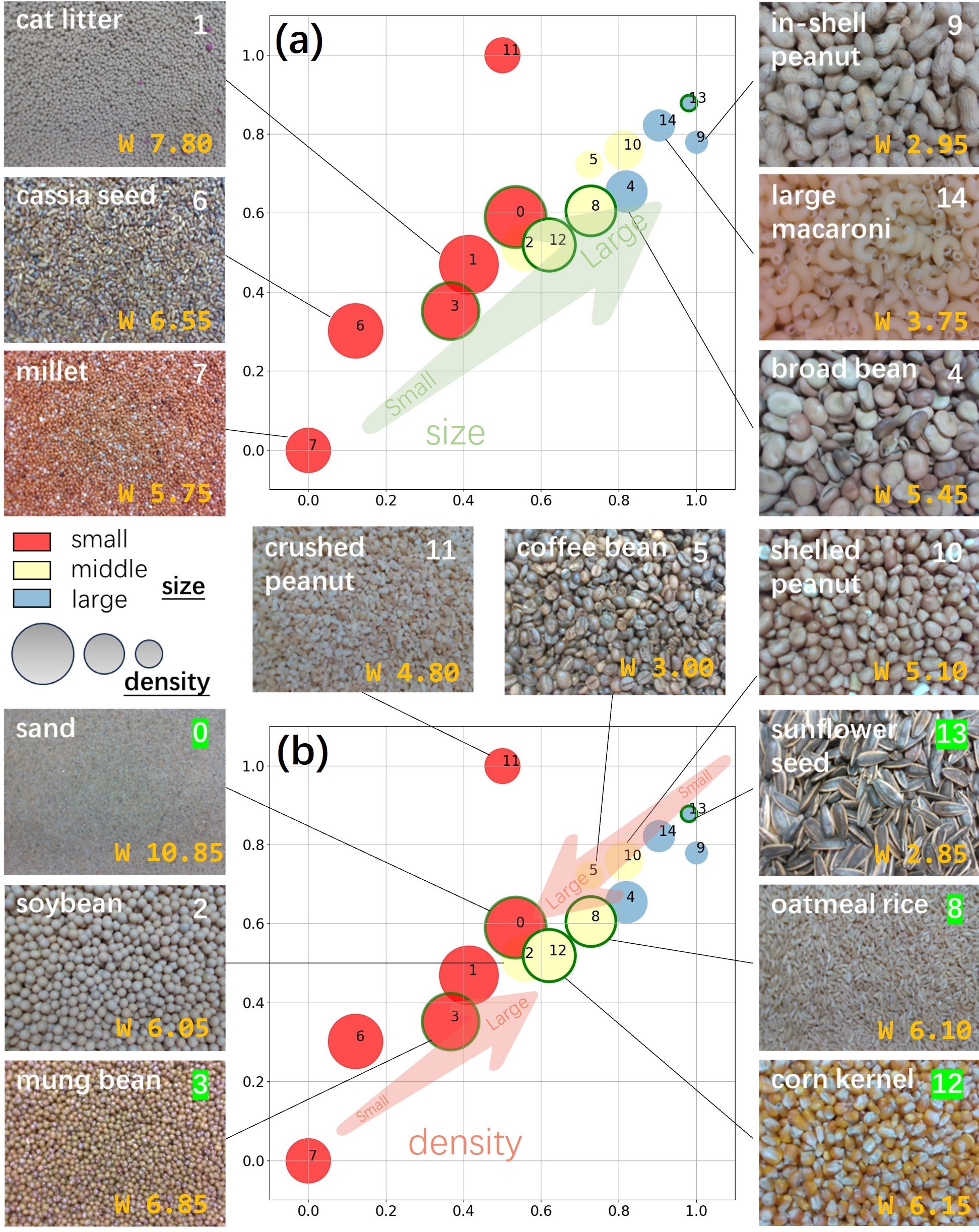}
    \vspace{-20pt}
    \caption{Implicit property distribution in two selected latent features. (a) GMs are arranged diagonally according to particle size. (b) GMs with large densities tend to be more concentrated in the latent embedding. Here, circle size refers to the value of particle density, determined by the weights of GMs in the same container, as shown at the bottom right of each GM display (unit: kg). The circle color indicates particle size according to the manual categorization. Note that these marker sizes and colors are only for the purpose of visualization. Also, the features of unseen GMs (with IDs in green) are outlined in green edges.}
    \label{fig:estimated_prop}
    \vspace{-20pt}
\end{figure}

In this paper, we perform visuo-haptic learning based on the contact model \prettyref{eq:third_law} from the study of granule media \cite{albert1999slow}. The good test results regarding the force prediction from the video encourage us to discover whether the trained model has learned about the physical properties of granules (especially for the visual and haptic attributes, i.e., particle size $d_c$ and density $\rho$ in \prettyref{eq:third_law}) in its latent space. 
In the test set, there are $10$ videos per GM. After the encoder, we normalize the latent features and calculate the average value in each latent dimension for every particle and then we find two dimensions from $4$D latent features that reveal the implicit property distribution of granules, as demonstrated in \prettyref{fig:estimated_prop}.
For ease of observation, we visualize the latent representations of particles according to their real physical properties. We broadly categorize GMs into three size ranges - large, medium, and small - and assign different colors to each size class. In addition, we utilize varying marker sizes to distinguish particulates on the basis of their density, determined by measuring the mass of the same volume.


In \prettyref{fig:estimated_prop}-(a), the particle size distribution exhibits a progression from small to large, with smaller particles like millet and cassia seed at the bottom-left, moderate sizes like coffee bean in the middle, and larger irregular shapes like large macaroni at the top-right.
In addition, from \prettyref{fig:estimated_prop}-(b), the particle density distribution exhibits a clustering pattern, with low-density large particles like in-shell peanut at the top-right, low-density small particles like millet at the bottom-left, and heavier granules like cat litter in the central region.

For unseen granules, their properties conform to the aforementioned distribution, as depicted with green edges in \prettyref{fig:estimated_prop}. For instance, sand has a small size but the heaviest weight, hence it aligns with the small size distribution and is positioned at the center of the density distribution as well. In addition, the sunflower seed has a large volume but a relatively light weight, placing them in the upper right corner of the implicit property distribution. 
However, the distribution of crushed peanuts deviates significantly from the mainstream pattern, likely due to the tendency of crushed peanuts to release oil, leading to particle adhesion and compromised data collection through poor particle tracking.
The above experimental findings not only enhance the interpretability of the model on the physical properties of GMs but also provide a particle property estimator using videos.

\begin{figure}
    \centering
    \includegraphics[width=0.485\textwidth]{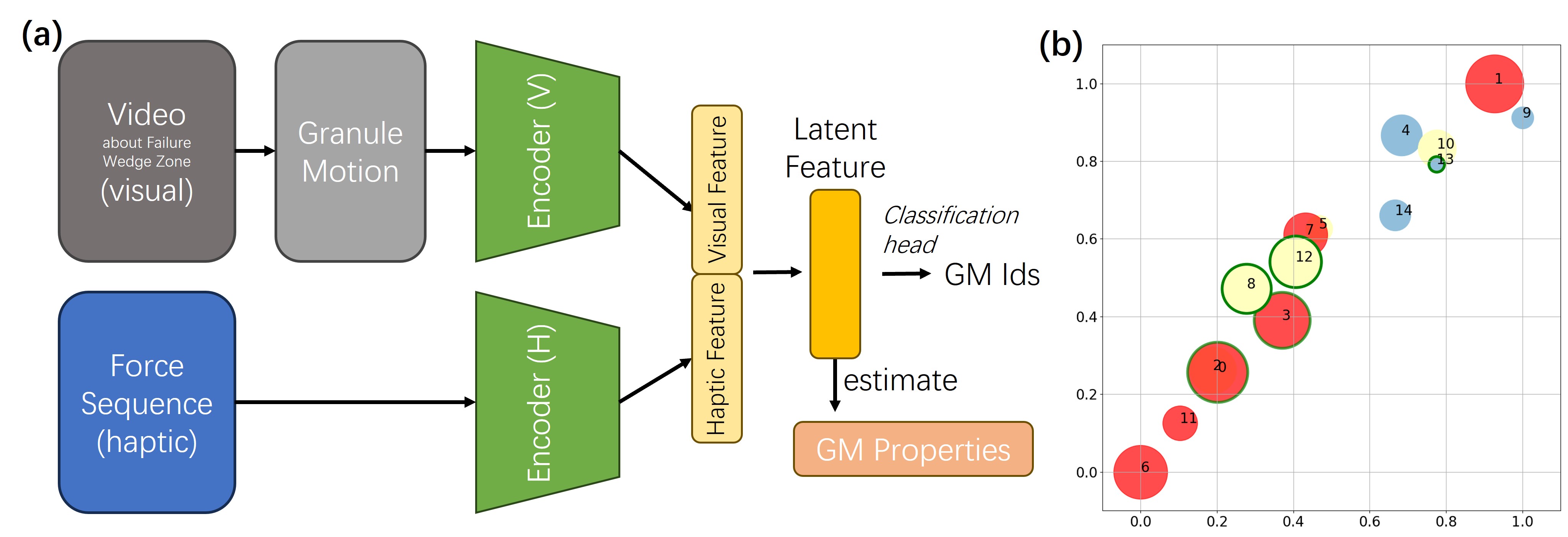}
    \vspace{-15pt}
    \caption{Baseline method. (a) Workflow of the traditional visuo-haptic learning for GM classification. (b) Latent features from a hidden layer before the classification head.}
    \label{fig:baseline}
    \vspace{-15pt}
\end{figure}

\subsection{Comparison with Baseline Method}
We compare our model with a baseline method based on a traditional visuo-haptic learning framework, where both video and force data are inputs, as exhibited in \prettyref{fig:baseline}-(a). We retain the particle tracking preprocessing and the encoder for the inputting granule motion is the same as our encoder, and the encoder for the force data is composed of four 1D convolution layers, which is the opposite of our decoder. Then, the concatenated visual and haptic features are mixed by a hidden layer, followed by a classification head for GM types via a softmax layer. The reasons for the classification head are as follows. For the supervised learning of particle properties, it is expected that the black-box model can learn a good relationship between visual-haptic data and particle size and density. However, this requires additional human effort to collect the true value of the GM attributes. Moreover, such a model is difficult to interpret, and the generalization is predictably poor. Therefore, in the same case where there is no true value of the physical properties, we propose the above model as a baseline method and try to explain the physical properties from its hidden layer.

\prettyref{fig:baseline}-(b) provides the resulting 2D visualization of the latent embeddings of the baseline method. From it, particles with different properties gather together, and
no interpretation pattern is found regarding the particle size and density. In addition, in the inference stage, both video and force data are needed as input for the baseline approach. However, for our method, only visual modality is required, which means we just need a camera to capture a probe-dragging video, and no force sensor is needed. This single-mode input greatly improves the usability of our method, especially in the field experiment, as demonstrated in \prettyref{sec:applications}.

\subsection{Evaluation of Particle Tracking Preprocessing}\label{sec:ablation}
\begin{figure}
    \centering
    \includegraphics[width=0.485\textwidth]{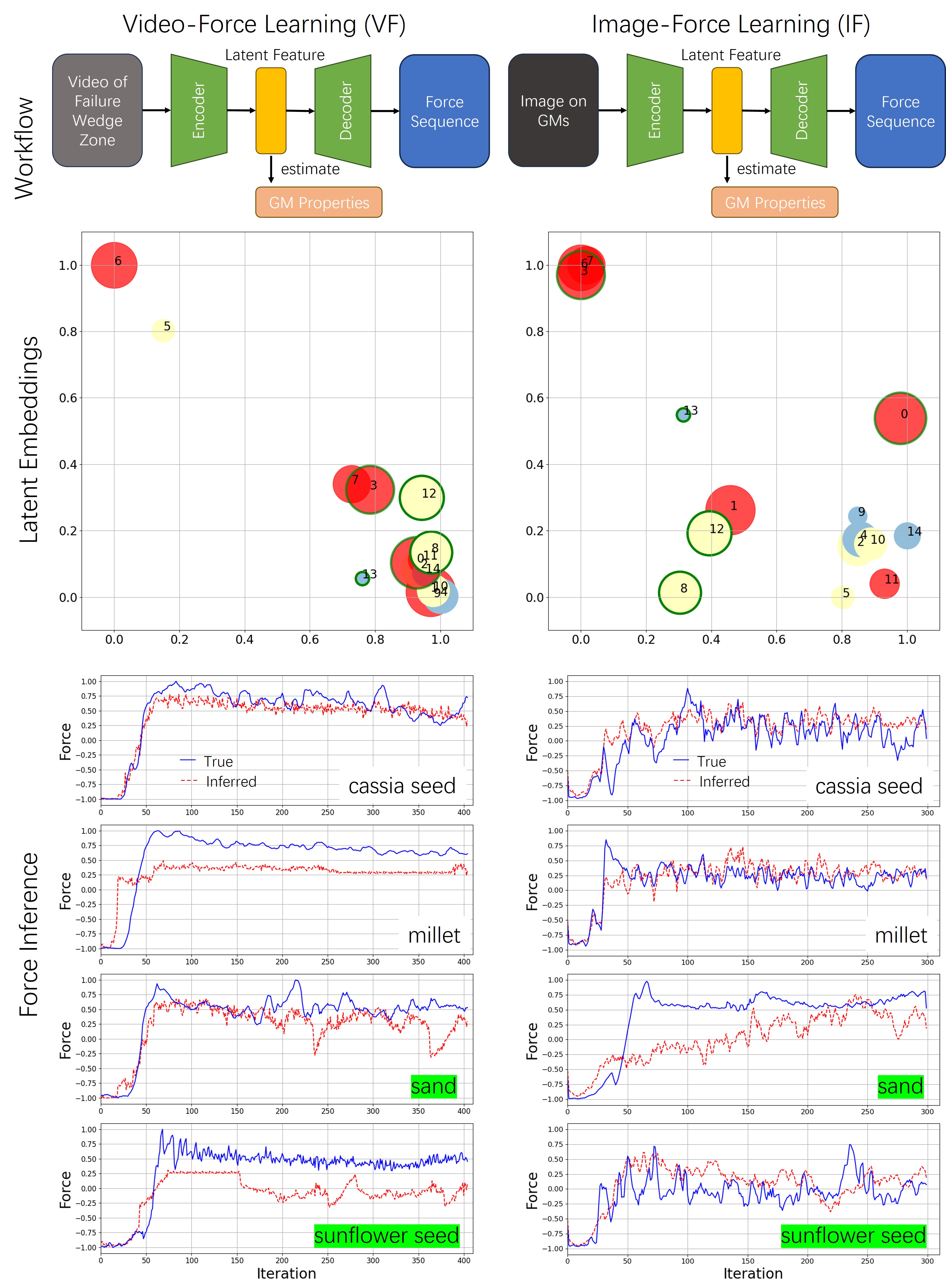}
    \vspace{-20pt}
    \caption{Ablation experiments trained on the video-force data and image-force data to evaluate the particle tracking preprocessing.}
    \label{fig:comp_exp_VF_IF}
    \vspace{-10pt}
\end{figure}

As depicted in \prettyref{fig:framework_mae}, we utilize the particle tracking algorithm to extract the granule motion information and further reduce the input dimension. To evaluate the particle tracking preprocessing, we also consider the ablation experiments of the proposed encoder-decoder network trained on the video-force data (denoted by VF method) and image-force data (denoted by IF method), respectively. The workflow of VF and IF methods are given in \prettyref{fig:comp_exp_VF_IF}. {Please refer to the project site given in the abstract for more details.} 

The 2D visualization of their latent embeddings and some force-inference examples are demonstrated in \prettyref{fig:comp_exp_VF_IF}. Compared to \prettyref{fig:estimated_prop}, we can observe that both the IF method and VF method exhibit scattered particle distributions in the latent features, making it difficult to differentiate granule properties based on the distribution. Alternatively, it is more prudent to state that the low-dimensional projection of the latent space fails to reflect the physical properties of the GM. 

For force inference analysis, compared to \prettyref{fig:force_inference}-(a), it can be seen that the VF method provides a relatively accurate estimation for cassia seed but exhibits larger prediction errors for other GMs. However, the results of the IF learning show acceptable performance on seen GMs but less favorable performance on unseen GMs. The above observations are consistent with the results of average loss calculation, as revealed in \prettyref{tab:ablation_exp}.
It can be found that the VF method, directly using videos without any preprocessing, has the highest loss. This seems to indicate that using particle tracking technology to extract the particle motion features explicitly is more instructive than allowing the model to learn the motion characteristics from videos by itself. Also, our method achieves a lower loss than the IF learning. This may suggest that the temporal visual signals on granules, as opposed to static images, are more favorable for model convergence to generate the time-series force data.

In addition, as expected, the VF learning method achieves the highest time cost for model training (around 2 h) and testing ($\sim 80$ms/case) in the same dataset, as reported in \prettyref{tab:ablation_exp}. Due to the proposed particle tracking technique, the computational burden of our method is significantly reduced, around 10 min for preprocessing and 30 min for training. Similarly, because of the low dimensions of the image compared to the video, the time cost of the IF approach is relatively small.

\begin{table}[!ht]
    \caption{Average Loss and Time Cost. Bold for lowest values.}
    \vspace{-5pt}
    \centering    
    \resizebox{0.43\textwidth}{!}{%
    \begin{tabular}{@{}ccccc@{}} 
    \toprule
        \multirow{2}{*}{Method} & \multicolumn{2}{c}{Avg Loss} & \multicolumn{2}{c}{Time Cost}\\ 
        \cmidrule(l){2-5} 
         & \multicolumn{1}{c}{Test Set} & \multicolumn{1}{c}{Unseen GMs} & \multicolumn{1}{c}{Train(min)} & \multicolumn{1}{c}{Test(ms/case)}
         \\\midrule\midrule
        IF   & 0.1431 & 0.1982 & $\sim 30 $& $\sim 2$ \\ \cmidrule(){1-5}
        VF   & 0.1711 & 0.2134 & $\sim 120$ & $\sim 80$   \\ \cmidrule(){1-5}
        Ours & \textbf{0.1368} & \textbf{0.1609} & $\sim 40$ & $\sim 2$ \\ 
    \bottomrule
    \end{tabular}
    } \vspace{-15pt}
    \label{tab:ablation_exp}
\end{table}

\subsection{Generalization Validation}
Based on the good interpretability of our model, its generalization capabilities will be verified in this subsection.


\subsubsection{Unseen GMs Collected by Robot Arm}
We have exhibited the generalization performance of our model on unseen GMs in previous subsections, as exhibited in \prettyref{fig:force_inference} and \prettyref{fig:estimated_prop}, respectively.

\subsubsection{Seen/Unseen GMs Collected by Handheld Device}
Thanks to the single-mode input in the visuo-haptic learning proposed in this paper, we only need videos without force signals in the model evaluation. 
So we can feed the videos recorded by the handheld device to the trained model and then try to analyze the physical properties of those particles involved in the videos. This aims to validate our model's generalization capability to data acquisition devices.  

\begin{figure}[!tb]
    \centering
    \includegraphics[width=0.48\textwidth]{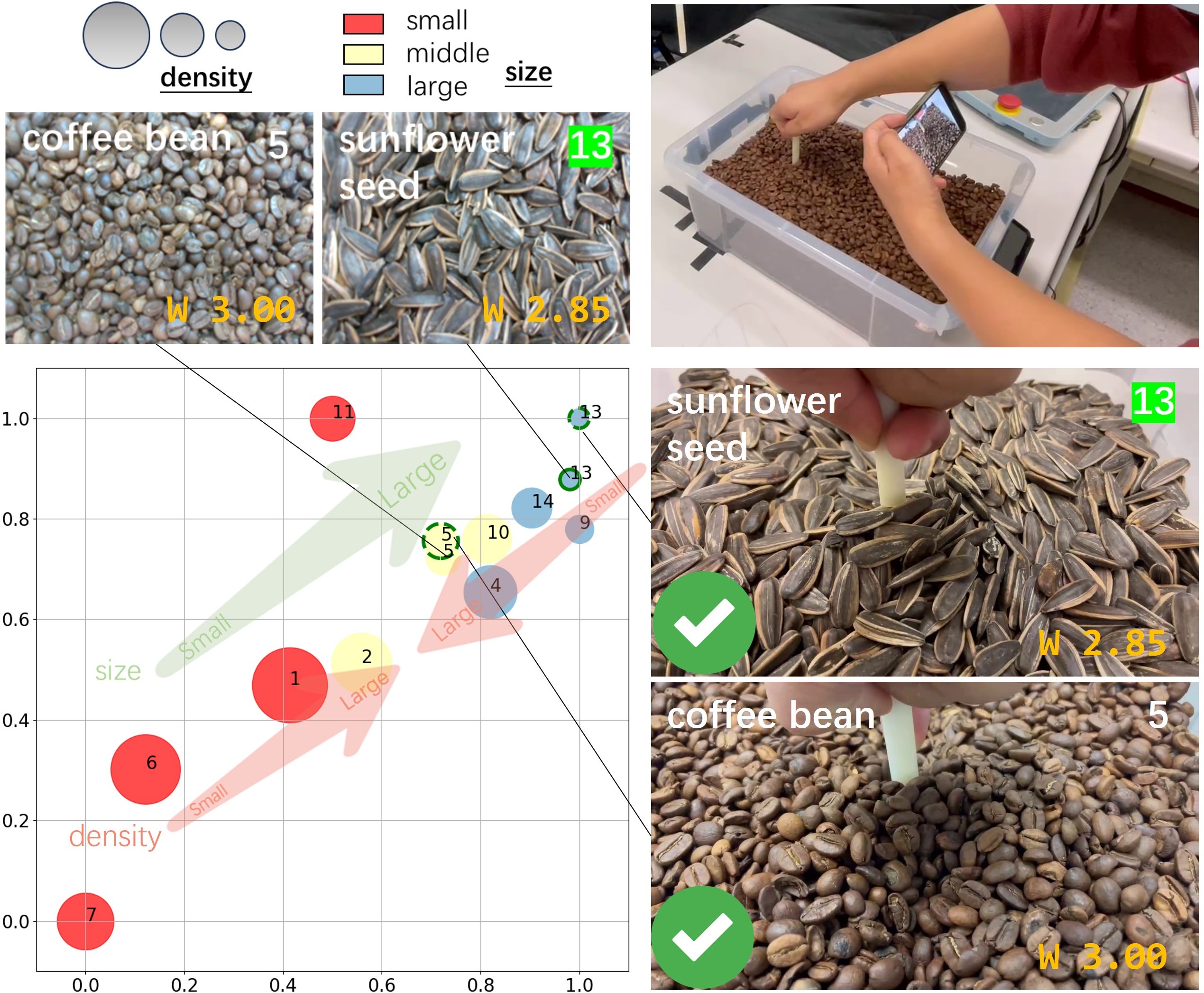}
    \vspace{-5pt}
    \caption{Video collection by the handheld device (top-right). We test on a seen GM, i.e., coffee bean, and an unseen particle, i.e., sunflower seed, whose property estimations are shown with the dashed green edge.}
    \label{fig:hand_hold_exp}
    \vspace{-15pt}
\end{figure}

In the same lab environment, as exhibited in \prettyref{fig:hand_hold_exp}, an operator holds the probe (with the same dimension as the one mounted on the robot arm) in the right hand and begins sliding it after randomly inserting granules. The operator's left-hand holds a smartphone, capturing the interaction between the probe and particles from a horizontal perspective. It is evident that compared to the robotic acquisition setup in \prettyref{fig:exp_setup_visuo_haptic_data}, the trajectory of the probe during manual collection is not perfectly straight, and the depth of probe insertion cannot remain consistent throughout the process. In addition, due to the dynamic relationship between the probe and the camera, the position of the probe in the video also varies. {We do not restrict the sliding speed and depth to be the same as those used during data collection by the robotic arm.} These factors introduce noise to the model's input and pose challenges to its generalization capabilities.

After sending videos to the trained encoder, we obtain their projections in the implicit property distribution, as presented in \prettyref{fig:hand_hold_exp}.
It is observed that, for the seen particle, coffee beans, the manually-recorded video exhibits a very close projection to the existing projection from robot-recorded videos in the implicit property distribution. Similarly, for unseen sunflower seeds, the data collected by handheld devices and the data collected using the UR5 are not far apart in the implicit property distribution. Moreover, the data collected using handheld devices adheres to the distribution characteristics of particle size and density mentioned in \prettyref{fig:estimated_prop}.
So, we can conclude that our model can be generalized to handheld devices. In this way, we can remove the limitation of the hardware configuration in \prettyref{fig:exp_setup_visuo_haptic_data} and use a mobile phone and a probe to estimate the properties of GM in practical applications, as demonstrated in the next subsection.



\begin{figure}[!tb]
    \centering
    \includegraphics[width=0.48\textwidth]{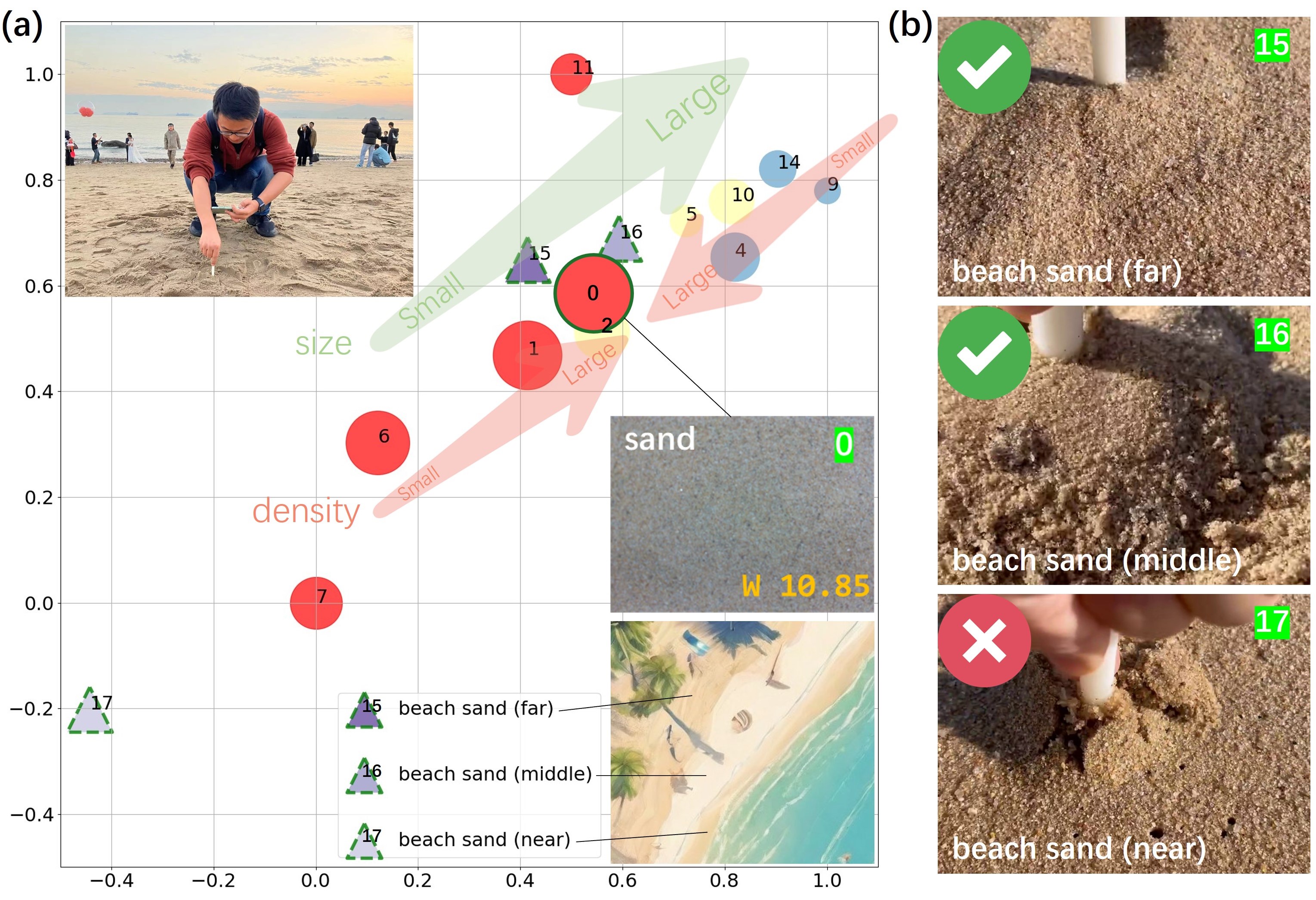}
    \vspace{-10pt}
    \caption{Application of property estimation for beach sands. We select three sampling sites according to the distance to the seawater and the snapshots of the probe-dragging are shown in (b). From (a), our model roughly gives its estimates of particle size and density for GM15 and GM16, which agree with the observations in the video. However, the model cannot be generalized to GM17, because the high water content in GM17 leads to the hardening of particles, which is different from the granule motion seen in the training.}
    \label{fig:beach_demo}
    \vspace{-15pt}
\end{figure}

\subsection{Property Estimation for Beach Sands}
\label{sec:applications}
We further extend the data collection scenario to the natural environment rather than the lab settings. Here, we use handheld devices to record the probe-dragging videos on a beach, as shown in the inset at the top left of \prettyref{fig:beach_demo}-(a). Specifically, we select three different sampling sites based on their proximity to seawater, and therefore these beach sands have different water content, denoted by beach sand (far), beach sand (middle) and beach sand (near), as depicted in the inset at the bottom right of \prettyref{fig:beach_demo}-(a). The water content in sands affects the failure wedge zone in front of the probe during the GM-probe interaction, as illustrated in snapshots in \prettyref{fig:beach_demo}-(b). It is obvious that these beach sands are entirely unknown to the model, resulting in newly assigned GM IDs with a green background in \prettyref{fig:beach_demo}-(b).

In the implicit property distribution, the model provides the estimated property distribution for these three unknown GMs, as shown in \prettyref{fig:beach_demo}-(a). We can observe that GM 15 and GM 16 are located near the center area, indicating that the model considers these two GMs to have relatively high density, which aligns with the observed property distribution of the sand from the dataset. Furthermore, the model assigns GM 15 and GM 16 with medium (or slightly large) particle sizes, which corresponds well to the aggregation of sand particles in the presence of water, as observed in the videos. However, the model's estimation for GM 17 exceeds the range of existing implicit property distribution. Upon inspection of the videos, we find that GM 17 has a significant moisture content due to its proximity to seawater. As a result, the particles in front of the probe exhibit large-scale cracking rather than the typical particle compression and stacking behavior in the traditional failure wedge zone. This deviation may render our model unable to generalize to this kind of GM. In contrast, GM 15 and GM 16 have relatively lower moisture content, and the granule movements in front of the probe are still similar to the observations exhibited in the lab. Thus, our model can generalize to these two cases well.


\subsection{Dimension of Latent Embedding}\label{sec:dim_latent}


{In this study, we use a 4D latent embedding, as shown in \prettyref{fig:framework_mae}. This choice comes from preliminary experiments with latent space dimensions ranging from 2 to 100, where we calculate MSE loss between predicted outputs and actual force sequences from the validation set. Results in \prettyref{tab:effect_dim_latent_space} show that the 4D latent space yields the smallest loss. We believe that for models with low MSE, GM properties may be well embedded in this latent space. After balancing the trade-off between performance and computational cost from high dimensions, we ultimately select 4 dimensions.}

\begin{table}[!ht]
    \caption{Average Loss in Validation Set for Different Dimensions of Latent Space. Bold for the lowest value.}
    \vspace{-5pt}
    \centering    
    \resizebox{0.4\textwidth}{!}{%
    \begin{tabular}{@{}cccccc@{}}
    \toprule
        Dim. & 2 & 3 & 4 & 5 & 6  \\ \cmidrule(){1-6}
        Avg Loss & 0.0170 & 0.0118 & \textbf{0.0104} & 0.0120 & 0.0114  \\ \cmidrule(){1-6}
        Dim. & 8 & 10 & 20 & 40 & 100  \\ \cmidrule(){1-6}
        Avg Loss & 0.0110 & 0.0112 & 0.0114 & 0.0113 & 0.0108 \\
    \bottomrule
    \end{tabular}
    }
    \vspace{-15pt}
    \label{tab:effect_dim_latent_space}
\end{table}

\section{Conclusion}
\label{sec:conclusion}
In this paper, we propose a useful method to estimate the relative distribution of particle properties merely from the video of GM-probe interaction. This method is trained on a visuo-haptic learning framework, mapping from the visual modality to the haptic modality. This architecture is guided by a contact model in the probe-dragging scenario \cite{albert1999slow} involving the strong correlation between the visual-haptic data and particle properties.
{Utilizing the interpretability of the learning process, we successfully uncover the implicit property distribution of GMs within the latent embeddings. Consequently, the relative values of the particle properties can be estimated based on their projection positions within this implicit property distribution, as determined by the trained visual encoder.}
The property analysis and generalization performance of the presented property estimator are extensively validated in the paper.

A limitation of this work is that its scope is restricted to homogeneous granules. This is because the contact model \cite{albert1999slow} underlying this study is specifically a model of the drag force experienced in a single, homogeneous particle, and its efficacy in a mixture of diverse particle types has not been verified. Moreover, the efficacy of particle tracking directly affects the final estimation results, as revealed by the failure cases of crushed peanuts in \prettyref{fig:estimated_prop}. Therefore, one future work may incorporate other perceptual modalities in the proposed visuo-haptic learning pipeline, such as the event signal from event camera \cite{xing2023target}, contact data from the tactile sensor \cite{chen2023polymer}, etc. {The investigation of probe trajectory in GMs is also an interesting problem.}

\bibliographystyle{IEEEtran}
\typeout{}
\bibliography{IEEEabrv,mybibfiles}

\begin{thebibliography}{10}
\providecommand{\url}[1]{#1}
\csname url@rmstyle\endcsname
\providecommand{\newblock}{\relax}
\providecommand{\bibinfo}[2]{#2}
\providecommand\BIBentrySTDinterwordspacing{\spaceskip=0pt\relax}
\providecommand\BIBentryALTinterwordstretchfactor{4}
\providecommand\BIBentryALTinterwordspacing{\spaceskip=\fontdimen2\font plus
\BIBentryALTinterwordstretchfactor\fontdimen3\font minus \fontdimen4\font\relax}
\providecommand\BIBforeignlanguage[2]{{%
\expandafter\ifx\csname l@#1\endcsname\relax
\typeout{** WARNING: IEEEtran.bst: No hyphenation pattern has been}%
\typeout{** loaded for the language `#1'. Using the pattern for}%
\typeout{** the default language instead.}%
\else
\language=\csname l@#1\endcsname
\fi
#2}}

\bibitem{zik1992mobility}
O.~Zik, J.~Stavans, and Y.~Rabin, ``Mobility of a sphere in vibrated granular media,'' \emph{EPL (Europhysics Letters)}, vol.~17, no.~4, p. 315, 1992.

\bibitem{zhang2024haptic}
Z.~Zhang, R.~Jia, Y.~Yan, R.~Han, S.~Lin, Q.~Jiang, L.~Zhang, and J.~Pan, ``A haptic-based proximity sensing system for buried object in granular material,'' \emph{arXiv preprint arXiv:2411.17083}, 2024.

\bibitem{yang2021assessment}
M.-D. Yang, Y.-C. Hsu, W.-C. Tseng, C.-Y. Lu, C.-Y. Yang, M.-H. Lai, and D.-H. Wu, ``Assessment of grain harvest moisture content using machine learning on smartphone images for optimal harvest timing,'' \emph{Sensors}, vol.~21, no.~17, p. 5875, 2021.

\bibitem{rasheed2022soil}
M.~W. Rasheed, J.~Tang, A.~Sarwar, S.~Shah, N.~Saddique, M.~U. Khan, M.~Imran~Khan, S.~Nawaz, R.~R. Shamshiri, M.~Aziz, \emph{et~al.}, ``Soil moisture measuring techniques and factors affecting the moisture dynamics: A comprehensive review,'' \emph{Sustainability}, vol.~14, no.~18, p. 11538, 2022.

\bibitem{castilla2023thermal}
R.~Castilla-Arquillo, A.~Mandow, C.~J. P{\'e}rez-del Pulgar, C.~{\'A}lvarez-Llamas, J.~M. Vadillo, and J.~Laserna, ``Thermal imagery for rover soil assessment using a multipurpose environmental chamber under simulated mars conditions,'' \emph{IEEE Transactions on Instrumentation and Measurement}, 2023.

\bibitem{zhong2024risk}
J.~Zhong, Q.~Li, J.~Zhang, P.~Luo, and W.~Zhu, ``Risk assessment of geological landslide hazards using d-insar and remote sensing,'' \emph{Remote Sensing}, vol.~16, no.~2, p. 345, 2024.

\bibitem{matl2020inferring}
C.~Matl, Y.~Narang, R.~Bajcsy, F.~Ramos, and D.~Fox, ``Inferring the material properties of granular media for robotic tasks,'' in \emph{2020 ieee international conference on robotics and automation (icra)}.\hskip 1em plus 0.5em minus 0.4em\relax IEEE, 2020, pp. 2770--2777.

\bibitem{guo2023estimating}
X.~Guo, H.-J. Huang, and W.~Yuan, ``Estimating properties of solid particles inside container using touch sensing,'' in \emph{2023 IEEE/RSJ International Conference on Intelligent Robots and Systems (IROS)}.\hskip 1em plus 0.5em minus 0.4em\relax IEEE, 2023, pp. 8985--8992.

\bibitem{albert1999slow}
R.~Albert, M.~Pfeifer, A.-L. Barab{\'a}si, and P.~Schiffer, ``Slow drag in a granular medium,'' \emph{Physical review letters}, vol.~82, no.~1, p. 205, 1999.

\bibitem{cotracker}
N.~Karaev, I.~Rocco, B.~Graham, N.~Neverova, A.~Vedaldi, and C.~Rupprecht, ``Cotracker: It is better to track together,'' \emph{arXiv preprint arXiv:2307.07635}, 2023.

\bibitem{yang2024one}
L.~Yang, L.~Yang, H.~Sun, Z.~Zhang, H.~He, F.~Wan, C.~Song, and J.~Pan, ``One fling to goal: Environment-aware dynamics for goal-conditioned fabric flinging,'' \emph{arXiv preprint arXiv:2406.14136}, 2024.

\bibitem{yang2020rigid}
L.~Yang, F.~Wan, H.~Wang, X.~Liu, Y.~Liu, J.~Pan, and C.~Song, ``Rigid-soft interactive learning for robust grasping,'' \emph{IEEE Robotics and Automation Letters}, vol.~5, no.~2, pp. 1720--1727, 2020.

\bibitem{liu2022robot}
J.~Liu, Y.~Chen, Z.~Dong, S.~Wang, S.~Calinon, M.~Li, and F.~Chen, ``Robot cooking with stir-fry: Bimanual non-prehensile manipulation of semi-fluid objects,'' \emph{IEEE Robotics and Automation Letters}, vol.~7, no.~2, pp. 5159--5166, 2022.

\bibitem{huang2022understanding}
H.-J. Huang, X.~Guo, and W.~Yuan, ``Understanding dynamic tactile sensing for liquid property estimation,'' \emph{arXiv preprint arXiv:2205.08771}, 2022.

\bibitem{niu2023goats}
Y.~Niu, S.~Jin, Z.~Zhang, J.~Zhu, D.~Zhao, and L.~Zhang, ``Goats: Goal sampling adaptation for scooping with curriculum reinforcement learning,'' \emph{arXiv preprint arXiv:2303.05193}, 2023.

\bibitem{takahashi2019deep}
K.~Takahashi and J.~Tan, ``Deep visuo-tactile learning: Estimation of tactile properties from images,'' in \emph{2019 International Conference on Robotics and Automation (ICRA)}.\hskip 1em plus 0.5em minus 0.4em\relax IEEE, 2019, pp. 8951--8957.

\bibitem{longhini2023edo}
A.~Longhini, M.~Moletta, A.~Reichlin, M.~C. Welle, D.~Held, Z.~Erickson, and D.~Kragic, ``Edo-net: Learning elastic properties of deformable objects from graph dynamics,'' in \emph{2023 IEEE International Conference on Robotics and Automation (ICRA)}.\hskip 1em plus 0.5em minus 0.4em\relax IEEE, 2023, pp. 3875--3881.

\bibitem{clarke2018learning}
S.~Clarke, T.~Rhodes, C.~G. Atkeson, and O.~Kroemer, ``Learning audio feedback for estimating amount and flow of granular material,'' \emph{Proceedings of Machine Learning Research}, vol.~87, 2018.

\bibitem{takahashi2021uncertainty}
K.~Takahashi, W.~Ko, A.~Ummadisingu, and S.-i. Maeda, ``Uncertainty-aware self-supervised target-mass grasping of granular foods,'' in \emph{2021 IEEE International Conference on Robotics and Automation (ICRA)}.\hskip 1em plus 0.5em minus 0.4em\relax IEEE, 2021, pp. 2620--2626.

\bibitem{kiyokawa2019generation}
T.~Kiyokawa, M.~Ding, G.~A.~G. Ricardez, J.~Takamatsu, and T.~Ogasawara, ``Generation of a tactile-based pouring motion using fingertip force sensors,'' in \emph{2019 IEEE/SICE International Symposium on System Integration (SII)}.\hskip 1em plus 0.5em minus 0.4em\relax IEEE, 2019, pp. 669--674.

\bibitem{kadokawa2023learning}
Y.~Kadokawa, M.~Hamaya, and K.~Tanaka, ``Learning robotic powder weighing from simulation for laboratory automation,'' in \emph{2023 IEEE/RSJ International Conference on Intelligent Robots and Systems (IROS)}.\hskip 1em plus 0.5em minus 0.4em\relax IEEE, 2023, pp. 2932--2939.

\bibitem{maladen2009undulatory}
R.~D. Maladen, Y.~Ding, C.~Li, and D.~I. Goldman, ``Undulatory swimming in sand: subsurface locomotion of the sandfish lizard,'' \emph{science}, vol. 325, no. 5938, pp. 314--318, 2009.

\bibitem{zhu2019data}
Y.~Zhu, L.~Abdulmajeid, and K.~Hauser, ``A data-driven approach for fast simulation of robot locomotion on granular media,'' in \emph{2019 international conference on robotics and automation (ICRA)}.\hskip 1em plus 0.5em minus 0.4em\relax IEEE, 2019, pp. 7653--7659.

\bibitem{kobayakawa2018local}
M.~Kobayakawa, S.~Miyai, T.~Tsuji, and T.~Tanaka, ``Local dilation and compaction of granular materials induced by plate drag,'' \emph{Physical Review E}, vol.~98, no.~5, p. 052907, 2018.

\bibitem{swick1988model}
W.~Swick and J.~Perumpral, ``A model for predicting soil-tool interaction,'' \emph{Journal of Terramechanics}, vol.~25, no.~1, pp. 43--56, 1988.

\bibitem{kingma2014adam}
D.~P. Kingma and J.~Ba, ``Adam: A method for stochastic optimization,'' \emph{arXiv preprint arXiv:1412.6980}, 2014.

\bibitem{xing2023target}
W.~Xing, S.~Lin, L.~Yang, and J.~Pan, ``Target-free extrinsic calibration of event-lidar dyad using edge correspondences,'' \emph{IEEE Robotics and Automation Letters}, 2023.

\bibitem{chen2023polymer}
W.~Chen, Y.~Yan, Z.~Zhang, L.~Yang, and J.~Pan, ``Polymer-based self-calibrated optical fiber tactile sensor,'' in \emph{2023 IEEE/RSJ International Conference on Intelligent Robots and Systems (IROS)}.\hskip 1em plus 0.5em minus 0.4em\relax IEEE, 2023, pp. 10\,197--10\,203.

\end{thebibliography}
\end{document}